\documentclass[letterpaper]{article} 
\usepackage{aaai2026}  
\usepackage{times}  
\usepackage{helvet}  
\usepackage{courier}  
\usepackage[hyphens]{url}  
\usepackage{graphicx} 
\usepackage{tabularray}
\usepackage{natbib} 
\usepackage{caption}
\usepackage{subcaption}

\usepackage{amsmath}
\usepackage{amssymb}
\usepackage{algorithm}
\usepackage{algpseudocode}

\title{StateSpace-SSL: Linear-Time Self-supervised Learning for \\ Plant Disease Detection}

\author{
    Abdullah Al Mamun\textsuperscript{\rm 1,\rm 2},
    Miaohua Zhang\textsuperscript{\rm 2},
    David Ahmedt-Aristizabal\textsuperscript{\rm 2}, \\
    Zeeshan Hayder\textsuperscript{\rm 2},
    Mohammad Awrangjeb\textsuperscript{\rm 1}
}

\affiliations{
    \textsuperscript{\rm 1}School of Information and Communication Technology, Griffith University, Nathan, Queensland 4111, Australia\\
    \textsuperscript{\rm 2}Imaging and Computer Vision Group, Data61, CSIRO, Black Mountain, Canberra 2601, Australia\\
    {\tt\small \{a.mamun,~m.awrangjeb\}@griffith.edu.au,}\\
    {\tt\small \{mam012,~miaohua.zhang,~david.ahmedtaristizabal,~zeeshan.hayder\}@data61.csiro.au}
}

\begin{document}

\maketitle

\begin{abstract}
Self-supervised learning (SSL) is attractive for plant disease detection as it can exploit large collections of unlabeled leaf images, yet most existing SSL methods are built on CNNs or vision transformers that are poorly matched to agricultural imagery. CNN-based SSL struggles to capture disease patterns that evolve continuously along leaf structures, while transformer-based SSL introduce quadratic attention cost from high-resolution patches. To address these limitations, we propose StateSpace-SSL, a linear-time SSL framework that employs a Vision Mamba state-space encoder to model long-range lesion continuity through directional scanning across the leaf surface. A prototype-driven teacher–student objective aligns representations across multiple views, encouraging stable and lesion-aware features from labelled data. Experiments on three publicly available plant disease datasets show that StateSpace-SSL consistently outperforms the CNN- and transformer-based SSL baselines in various evaluation metrics.  Qualitative analyses further confirm that it learns compact, lesion-focused feature maps, highlighting the advantage of linear state-space modelling for self-supervised plant disease representation learning.

\end{abstract}

\section{Introduction}

Plant diseases are a persistent threat to global food security, reducing crop productivity and undermining agricultural sustainability across diverse farming systems \cite{al2024plant}. Early and reliable disease detection is therefore essential for stable agricultural production. While deep learning has advanced automated plant diagnosis, supervised pipelines rely on large annotated datasets that are costly to create, difficult to maintain across species and seasons, and often impractical under diverse field conditions \cite{barbedo2019plant}. These limitations have increased interest in self-supervised learning (SSL), which can learn leaf-level representations directly from unlabelled images.

\begin{figure}[!t]
    \centering
    \includegraphics[width=0.8\linewidth]{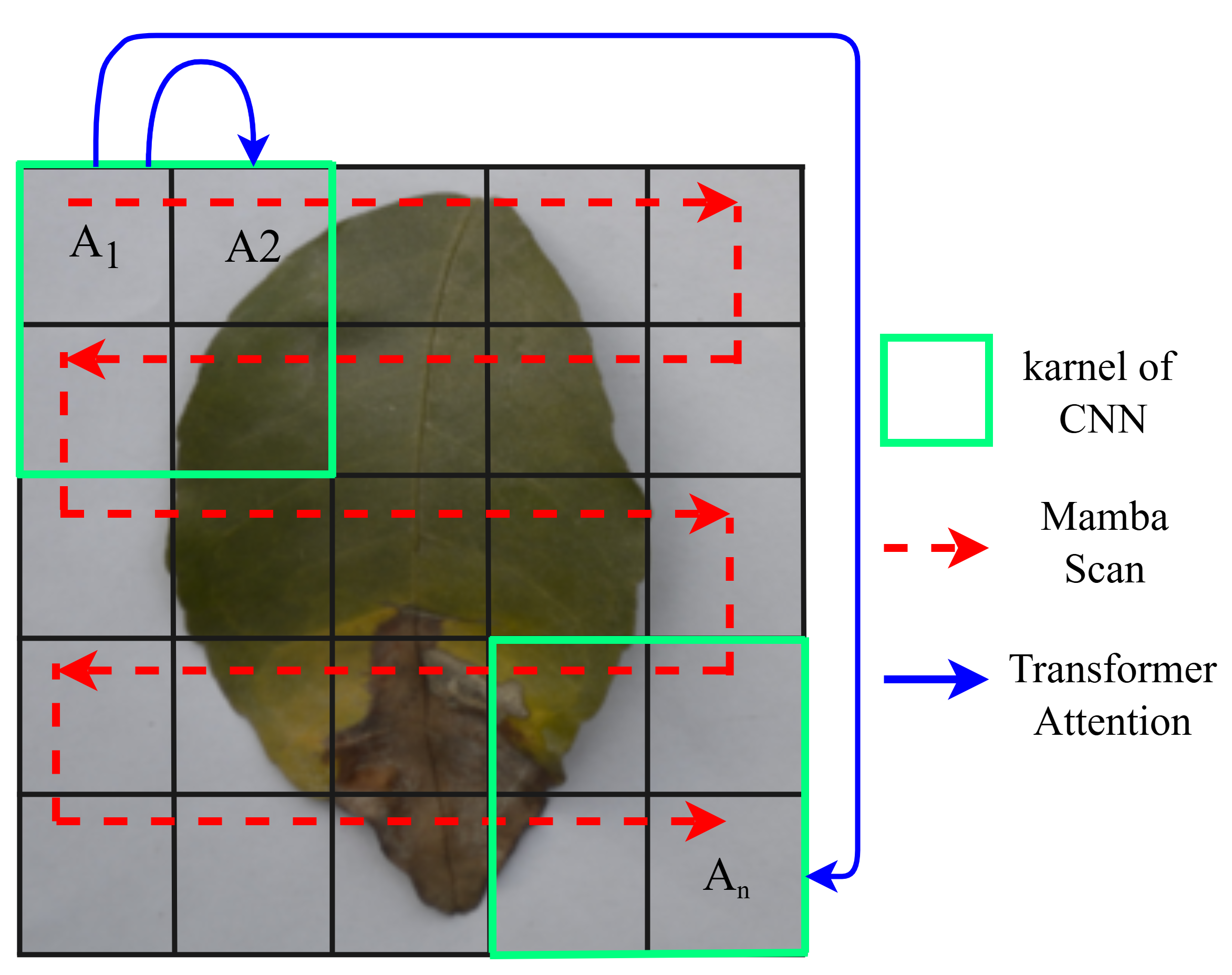}
    \caption{Spatial modelling for plant disease detection: CNN (green), Transformer (blue), and Mamba state-space scanning (red).}
    \label{fig:scan}
\end{figure}

Recent SSL frameworks span several paradigms. Early approaches such as SimCLR \cite{chen2020simple}, MoCo \cite{chen2020improved}, BYOL \cite{grill2020bootstrap}, SimSiam \cite{chen2021exploring}, SwAV \cite{caron2020unsupervised}, and DeepCluster-v2 \cite{caron2018deep} are primarily built on convolutional neural network (CNN) backbones. Although effective for natural images, CNN-based SSL is restricted by local receptive fields and struggles to capture the long-range spatial progression of plant diseases, where lesions evolve across the leaf surface along continuous structural patterns. More recent SSL variants, including MAE \cite{he2022masked}, Vision Transformer (ViT) \cite{dosovitskiy2020image}, DINO \cite{caron2021emerging}, and BEiT \cite{bao2021beit}, adopt transformers to incorporate global context. However, self-attention in Transformers introduces quadratic complexity with respect to token count. High-resolution leaf images, combined with the multi-crop strategy required in SSL, amplify this cost and make transformer-based SSL memory-intensive and difficult to scale in agricultural settings. Together, these limitations highlight the need for a representation learning architecture that can model both local and long-range dependencies while remaining computationally efficient.

\begin{figure*}[!t]
    \centering
    \includegraphics[width=0.9\linewidth]{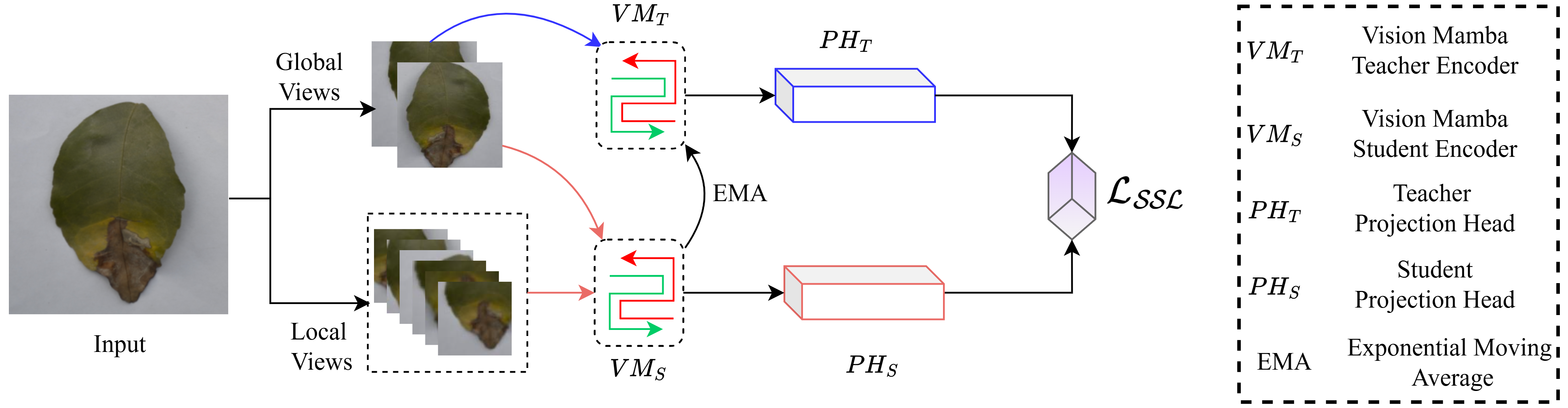}
    \caption{Schematic diagram of the proposed StateSpace-SSL framework, integrating Vision Mamba encoders with prototype-based teacher–student alignment across global and local views.}
    \label{fig:proposed_method}
\end{figure*}

Beyond architectural constraints, plant disease imagery presents distinct spatial and structural challenges. Lesion symptoms evolve in continuous patterns along the leaf surface. As illustrated in Figure \ref{fig:scan}, CNNs captures only localised textures and cannot account for disease progression across wider spatial regions. Transformers capture global relations but incur quadratic complexity with respect to token count, and often produce diffuse attention patterns in field environments with clutter and noise. As a result, neither architecture naturally reflects how disease symptoms propagate spatially, limiting their effectiveness in self-supervised plant disease detection.

State-space models (SSMs) offer a compelling alternative. Unlike CNN kernels or quadratic attention layers, SSMs such as VM \cite{zhu2024vision} capture structured long-range dependencies using a directional and recurrent scanning mechanism that scales linearly with sequence length. This behaviour mirrors the spatial continuity of plant diseases, where disease severity changes progressively across the leaf. As a result, SSMs provide an inductive bias that enables the extraction of fine-grained lesion cues and broader structural context with minimal computational cost.

Motivated by these observations, we introduce \textbf{StateSpace-SSL}, a linear-time self-supervised framework designed specifically for plant disease detection. The method replaces attention with a VM encoder that models spatial lesion dynamics through state-space recurrence, and learns stable disease-aware representations through prototype-based teacher–student alignment across multi-scale global and local views. This formulation enables efficient pretraining without labelled data while preserving the expressive capacity needed for plant disease recognition.

Our main contributions are summarised as follows:
\begin{itemize}
    \item We propose StateSpace-SSL, a novel self-supervised learning framework that leverages a Vision Mamba encoder to capture long-range dependencies in plant leaf imagery with linear-time complexity, offering an efficient solution for training on high-resolution agricultural images.
    \item We introduce a prototype-based teacher-student self-distillation strategy that aligns global and local views via shared prototypes, enabling stable and lesion-aware representation learning without quadratic computational cost. 
    \item We show that combining state-space modelling, multi-crop self-supervision, and prototype alignment produces a lightweight architecture that balances accuracy, robustness, and efficiency, making it suitable for real-world agricultural deployment. 
\end{itemize}

\section{Related Work}
SSL has become a key strategy for plant disease detection due to its ability to learn representations without costly annotations. Early SSL methods were built on CNN backbones such as SimCLR \cite{chen2020simple}, MoCo v2 \cite{chen2020improved}, BYOL \cite{grill2020bootstrap}, and SimSiam \cite{chen2021exploring}, which rely on contrasting augmented views or predicting momentum-encoded targets. While effective on natural images, convolutional SSL is limited by local receptive fields that struggle to represent the long-range spatial evolution of plant diseases, where lesions propagate along vein structures and interact with global leaf morphology. Approaches such as MaskCOV \cite{yu2021maskcov} have introduced covariance prediction and patch masking to enhance discriminability, yet CNNs inherently lack global context modelling and perform inconsistently under complex field environments.

Transformer-based SSL has extended these capabilities by introducing global self-attention. MAE \cite{he2022masked} learns a holistic structure via masked patch reconstruction with an asymmetric encoder–decoder, while DeiT \cite{touvron2021training} introduces data-efficient training and a distillation token for stable self-distillation. DINO \cite{caron2021emerging} employs a multi-crop teacher–student framework that produces strong semantic attention maps, and TransFG \cite{he2022transfg} adapts the ViT for fine-grained recognition using token refinement and region alignment. Hybrid formulations such as the Hybrid ViT (ViT combined with ResNet50) \cite{dosovitskiy2020image} fuse convolutional locality with transformer globality to improve fine-grained discrimination. Although powerful, these transformer-based SSL methods share a critical limitation: self-attention scales quadratically with token count. High-resolution leaf images and multi-crop SSL augmentations dramatically inflate token lengths, resulting in high memory consumption, long training cycles, and limited scalability. Moreover, transformers can generate diffuse or background-sensitive attention in field images, lacking an inductive bias aligned with the spatial continuity of disease progression.

These limitations indicate the need for an SSL framework that can capture long-range lesion continuity while remaining computationally scalable for high-resolution agricultural imagery. State-space models (SSMs), particularly Mamba-based architectures \cite{gu2023mamba}, offer an attractive alternative by modelling long-range dependencies through linear-time selective recurrence. VM \cite{zhu2024vision} processes patch sequences directionally, capturing spatial continuity without quadratic interactions and providing an inductive bias that is well-aligned with disease progression patterns. Despite these advantages, SSMs have not been explored within self-supervised plant disease detection.

These observations motivate StateSpace-SSL, which integrates VM with a prototype-based teacher–student SSL pipeline to learn efficient, lesion-aware representations. By avoiding attention’s quadratic bottleneck while retaining strong global modelling capacity, StateSpace-SSL directly addresses the structural and computational limitations of existing SSL methods.

\section{Method}

\subsection{Overview}
StateSpace-SSL is an SSL framework that combines a linear-time VM encoder with a teacher-student self-distillation strategy. Given an unlabelled input image $I$, we generate multi-crop views and process them through a student encoder $f_{\theta}$ and a teacher encoder $f_{\xi}$ updated via an Exponential Moving Average (EMA). Training optimises a distributional alignment objective in which the student matches the teacher’s prototype predictions across multi-scale global and local views. This enables the model to learn disease-relevant spatial patterns while incurring substantially lower computational overhead than attention-based SSL. An overview of the proposed framework is provided in Figure \ref{fig:proposed_method}.

\subsection{State-space Encoder}
Transformers capture long-range dependencies with quadratic token complexity, which is costly for high-resolution plant images. VM replaces full-token attention with a linear-time state-space recurrence, providing an efficient yet expressive alternative.

The continuous-time state-space formulation is
\begin{equation}
  \frac{d h(t)}{d t} = A h(t) + B u(t), 
  \qquad 
  y(t) = C h(t),
\end{equation}
where $u(t)$ is the input token, $h(t)$ the latent state, and $y(t)$ the output.
VM implements a discretised update of the form
\begin{equation}
  h_k = g_k \odot (W_s h_{k-1} + W_x x_k),
\end{equation}
where $x_k$ is the $k$-th patch token, $h_k$ the updated state, $W_s$ and $W_x$ are learnable matrices, and $g_k$ is a learned gate. This recurrence scales as $\mathcal{O}(N)$ in sequence length $N$.

Given an image $I \in \mathbb{R}^{H \times W \times 3}$, we extract patch embeddings
\begin{equation}
  X = \{x_1, \dots, x_N\}, 
  \qquad 
  N = \frac{H W}{p^{2}},
\end{equation}
for a patch size $p$. The VM encoder processes this sequence through stacked Mamba blocks and outputs token features and hidden-state trajectories
\begin{equation}
  (Z, S) = f_{\theta}(X),
\end{equation}
where $Z = \{z_i\}$ are the final token embeddings and $S = \{h_i\}$ is the corresponding state sequence. This linear-time recurrence naturally captures spatially continuous lesion progression while remaining memory-efficient during SSL pretraining.

\begin{algorithm}[!t]
  \caption{Training pipeline of StateSpace-SSL}
  \label{alg:statespace_ssl}
  \begin{algorithmic}[1]
    \State \textbf{Input:} Unlabelled images $\{I_i\}$; initial student parameters $\theta$; teacher parameters $\xi \leftarrow \theta$; momentum $m$
    \State \textbf{Output:} Pretrained Vision Mamba encoder $f_{\theta}^\star$
    \For{each minibatch $\mathcal{B} = \{I_i\}$}
      \State Generate 2 global crops $v_1^{g}, v_2^{g}$ and 6 local crops $v_1^{\ell}, \dots, v_6^{\ell}$ for each $I_i$
      \State Assign teacher view $v^{g}_{t} = v_1^{g}$ and the student global view $v^{g}_{s} = v_2^{g}$
      \State Student forward pass:
      \[
        p_s = f^{\mathrm{Mamba}}_{\theta}
        \big(\{v^{g}_{s}, v_1^{\ell}, \dots, v_6^{\ell}\}\big)
      \]
      \State Teacher forward pass (single global view):
      \[
        p_t = f^{\mathrm{Mamba}}_{\xi}\big(v^{g}_{t}\big)
      \]
      \State Compute loss
      \vspace{-10pt}
      \[
        \mathcal{L}_{\mathrm{SSL}} = - \sum_{i=1}^{K} p_t(i)\log p_s(i)
      \]
      \State Update student parameters $\theta$ using AdamW with respect to $\mathcal{L}_{\mathrm{SSL}}$
      \State Update teacher via EMA: $\xi \leftarrow m \, \xi + (1-m)\theta$
    \EndFor
    \State \Return $f_{\theta}^\star$
  \end{algorithmic}
\end{algorithm}

\subsection{StateSpace-SSL Pipeline}

For each input image $I$, we generate two global crops $v_1^{g}, v_2^{g} \in \mathbb{R}^{224 \times 224}$ and six local crops $v_1^{\ell}, \dots, v_6^{\ell} \in \mathbb{R}^{96 \times 96}$. The student VM encoder $f_{\theta}$ processes one global view and six local views (seven views in total),, whereas the teacher encoder $f_{\xi}$ receives only two global views to provide stable targets.

Both encoders are equipped with a lightweight two-layer MLP projection head
\begin{equation}
  h(z) = W_2 \, \sigma ( W_1 z ),
\end{equation}
and produce prototype distributions via temperature-scaled softmax:
\begin{align}
  p_s &= \mathrm{softmax}\!\big(h_s(f_{\theta}(v)) / T_s\big), \\
  p_t &= \mathrm{softmax}\!\big(h_t(f_{\xi}(v')) / T_t\big),
\end{align}
where $T_s$ and $T_t$ are temperatures and $h_s, h_t$ denote the student and teacher heads, respectively. Teacher parameters are updated by EMA.
\begin{equation}
  \xi \leftarrow m \, \xi + (1-m)\theta,
\end{equation}
with momentum $m$ scheduled to increase during training.

The learning objective is a prototype-based distributional alignment loss that encourages consistency between teacher and student predictions:
\begin{equation}
  \mathcal{L}_{\mathrm{SSL}} 
  = - \sum_{i=1}^{K} p_t(i) \log p_s(i),
\end{equation}
where $K$ is the number of prototypes. This objective aligns the student with the teacher across both global and local views, promoting multi-scale feature consistency.

This self-supervised pipeline encourages the VM encoder to learn spatially coherent, disease-specific representations while preserving the efficiency benefits of linear-time state-space modelling. The overall training loop is summarised in Algorithm~\ref{alg:statespace_ssl}.

\begin{table}
\centering

\begin{tblr}{
cells = {c,m}, 
   rows = {font=\footnotesize},
  row{odd} = {c},
  row{4} = {c},
  row{6} = {c},
  row{8} = {c},
  row{10} = {c},
  row{12} = {c},
  row{14} = {c},
  cell{1}{1} = {r=2}{},
  cell{1}{2} = {r=2}{},
  cell{1}{3} = {c=3}{},
  hline{1,3,16} = {-}{},
  hline{2} = {3-5}{},
}
\textbf{Method}    & \textbf{Bb}      & \textbf{Accuracy}     &                       &                           \\
                   &                  & \textbf{\textbf{PV~}} & \textbf{\textbf{PD~}} & \textbf{\textbf{Citrus~}} \\
SimCLR (L)          & RN50             & 86.4                  & 78.8                  & 84.3                      \\
SimCLR (FT)        & RN50             & 88.7                  & 81.4                  & 86.1                      \\
MoCo v2 (L)        & RN50             & 85.2                  & 76.0                  & 83.1                      \\
MoCo v2 (FT)       & RN50             & 89.2                  & 82.1                  & 87.0                      \\
BYOL (L)           & RN50             & 88.5                  & 80.0                  & 85.4                      \\
BYOL (FT)          & RN50             & 90.0                  & 83.5                  & 87.2                      \\
MaskCOV            & RN50             & 92.0                  & 86.6                  & 89.6                      \\
DeiT               & ViT-S            & 92.2                  & 87.2                  & 89.0                      \\
TransFG            & ViT-S            & 92.8                  & 88.0                  & 89.5                      \\
MAE                & ViT-B            & 92.9                  & 88.1                  & 89.1                      \\
Hybrid ViT         & {ViT-S + \\RN50} & 93.0                  & 88.3                  & 89.7                      \\
DINO               & ViT-S            & 93.1                  & 88.6                  & 89.4                      \\
{StateSpace-SSL} & \textbf{VM}      & \textbf{94.61}        & \textbf{91.24}        & \textbf{89.83}  \\          
\end{tblr} \\
\scriptsize{\textbf{Note:} Bb= Backbone; PV= PlantVillage; PD= PlantDoc; VM= Vision Mamba; RN= ResNet; ViT= Vision Transformer; L= Large; FT= Fine Tune.}
\caption{Accuracy comparison across three plant disease datasets for different self-supervised learning models.}
\label{tab:accuracy_compare}
\end{table}

\section{Results and Discussion}

This section presents a comprehensive evaluation of StateSpace-SSL across multiple plant disease benchmarks, focusing on both recognition accuracy and computational efficiency. We first describe the experimental setup and datasets, followed by quantitative comparisons in terms of accuracy, parameter count, training time, and memory usage. We further analyse the per-epoch training behaviour and provide qualitative visualisations to assess the localisation quality of the learned representations. Together, these results demonstrate the effectiveness and practicality of the proposed state-space formulation for plant disease detection.

\subsection{Experimental Setup }

All experiments were performed on a multi-GPU compute node equipped with $10\times$ NVIDIA GPUs (11\,GB each). The models were implemented in PyTorch 2.1, utilising Mamba-SSM v2.2.3 for the state-space layers within the VM encoder. Distributed training was carried out using DistributedDataParallel with NCCL as the communication backend. The software environment included CUDA 11.8, Python 3.10, and Ubuntu 20.04.

\begin{table}
\centering
\begin{tblr}{
  cells = {c},
   rows = {font=\footnotesize},
  hline{1-2,14-15} = {-}{},
}
\textbf{Method} & \textbf{Backbone} & \textbf{PM}  & \textbf{CS} \\
MAE             & ViT-B             & 86M          & QT          \\
Hybrid ViT      & ViT-S + RN50      & 45M          & QT          \\
SimCLR (L)      & RN50              & 23M          & LP          \\
SimCLR (FT)     & RN50              & 23M          & LP          \\
MoCo v2 (L)     & RN50              & 23M          & LP          \\
MoCo v2 (FT)    & RN50              & 23M          & LP          \\
BYOL (L)        & RN50              & 23M          & LP          \\
BYOL (FT)       & RN50              & 23M          & LP          \\
MaskCOV         & RN50              & 23M          & LP          \\
DeiT            & ViT-S             & 22M          & QT          \\
TransFG         & ViT-S             & 22M          & QT          \\
DINO            & ViT-S             & 22M          & QT          \\
StateSpace-SSL  & VM                & \textbf{19M} & \textbf{LT} \\
\end{tblr} \\
\scriptsize{\textbf{Note:} VM= Vision Mamba; RN= ResNet; ViT= Vision Transformer; CS= Compute Scaling; QT= Quadratic in Tokens; LT= Linear in Pixels; LT= Linear in Tokens.}
\caption{Comparison of parameters and compute scaling across self-supervised learning models.}
\label{tab:parameters_compare}
\end{table}

\subsection{Datasets}

We evaluate StateSpace-SSL on three publicly available plant disease datasets covering both controlled and real-field conditions.
\textit{PlantVillage} \cite{hughes2015open} provides over 54,000 clean laboratory leaf images across 38 classes, serving as an in-domain benchmark.
\textit{PlantDoc} \cite{singh2020plantdoc} contains 2,598 field images spanning 28 diseases, with substantial variation in lighting, background clutter, and leaf morphology, making it suitable for cross-domain evaluation.
The \textit{Citrus} dataset \cite{rauf2019citrus} includes 609 high-resolution images from five disease types, enabling assessment of fine-grained lesion discrimination.

\subsection{Main Results}
StateSpace-SSL achieves consistently strong performance across all three benchmarks while requiring substantially fewer computational resources than competing methods. As shown in Table \ref{tab:accuracy_compare}, the model attains the highest accuracy on PlantVillage, PlantDoc, and Citrus, despite using a markedly smaller parameters count than the SSL baselines. The gains are especially pronounced on PlantDoc, where complex field conditions typically degrade the performance of attention-driven models.

\begin{table}
\centering
\begin{tblr}{
  cells = {c},
  rows = {font=\footnotesize},
  hline{1-2,8} = {-}{},
}
\textbf{Method} & \textbf{Backbone} & \textbf{Time} & \textbf{VRAM}  \\
MAE             & ViT-B             & 30h           & 22.6GB         \\
Hybrid ViT      & ViT-S + RN50      & 24h           & 19.8GB         \\
DINO            & ViT-S             & 20h           & 17.8GB         \\
TransFG         & ViT-S             & 18h           & 17.0GB         \\
DeiT            & ViT-S             & 17h           & 16.5GB         \\
StateSpace-SSL  & VM                & \textbf{9h}   & \textbf{8.7GB} 
\end{tblr}

\scriptsize{\textbf{Note:} VM= Vision Mamba; ViT= Vision Transformer.}
\caption{Training time and memory usage for different transformer-based self-supervised methods.}
\label{tab:timing_compare}
\end{table}

The parameter and complexity comparison in Table \ref{tab:parameters_compare} provides further context for this behaviour. Beyond differences in parameter count, the Compute Scaling (CS) highlights a fundamental contrast in architectural behaviour: most transformer-based baselines exhibit quadratic scaling with respect to tokens (QT), causing their computational cost to grow sharply with higher input resolutions or longer token sequences. 
StateSpace-SSL, by contrast, adopts linear scaling with respect to tokens  (LT), ensuring that compute increases proportionally rather than explosively as input size grows. Combined with its compact 19M-parameter VM backbone, this linear scaling makes the model substantially lighter and more tractable than transformer counterparts that typically exceed 45M–80M parameters. The resulting reduction in both model size and compute growth not only simplifies optimisation during self-supervised pretraining but also enables more efficient processing of diverse augmented views. This distinction in architectural complexity directly influences the favourable training characteristics observed in practice.

\begin{figure}[!t]
    \centering
    \includegraphics[width=.9\linewidth]{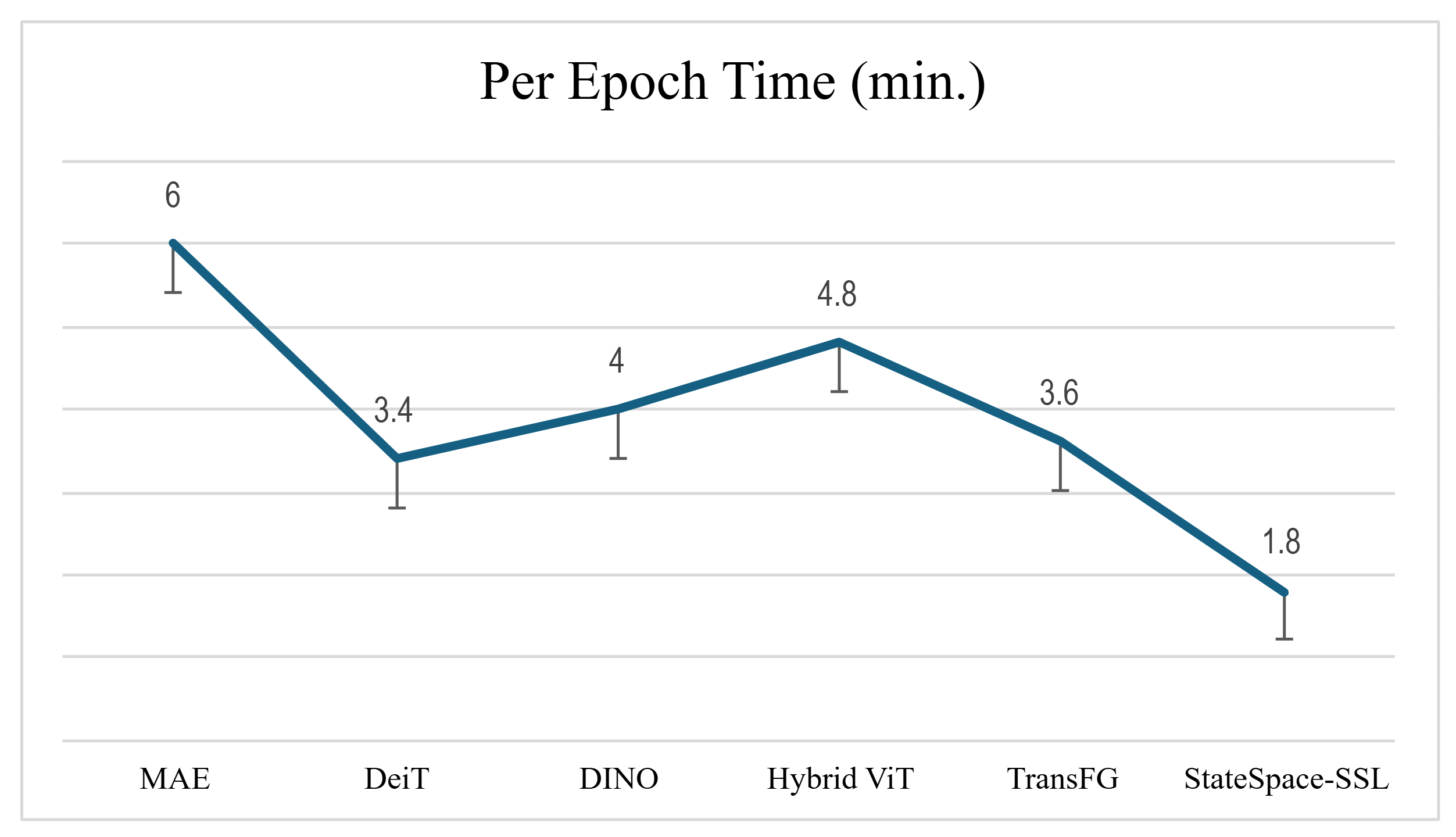}
    \caption{Per-epoch training time for different transformer-based self-supervised models.}
    \label{fig:time_per_epoch}
\end{figure}

\begin{figure}[!t]
    \centering

    \begin{subfigure}{0.12\textwidth}
        \centering
        \includegraphics[width=\linewidth]{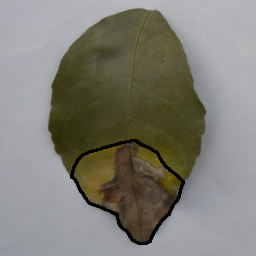}
        \caption{Input}
        \label{fig:a}
    \end{subfigure}
    \hspace{1.5em}
    \begin{subfigure}{0.12\textwidth}
        \centering
        \includegraphics[width=\linewidth]{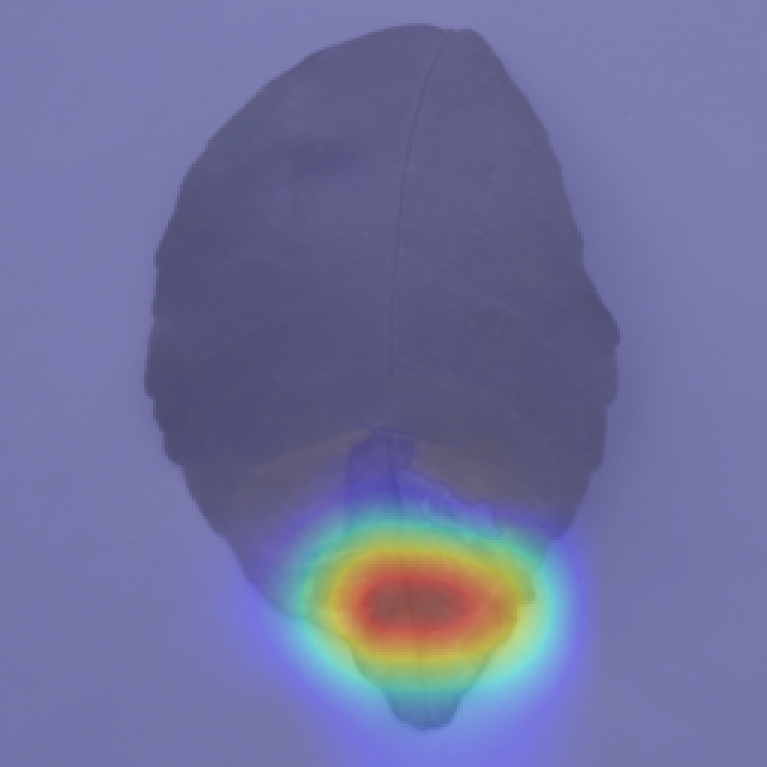}
        \caption{DINO}
        \label{fig:b}
    \end{subfigure}
    \hspace{1.5em}
    \begin{subfigure}{0.12\textwidth}
        \centering
        \includegraphics[width=\linewidth]{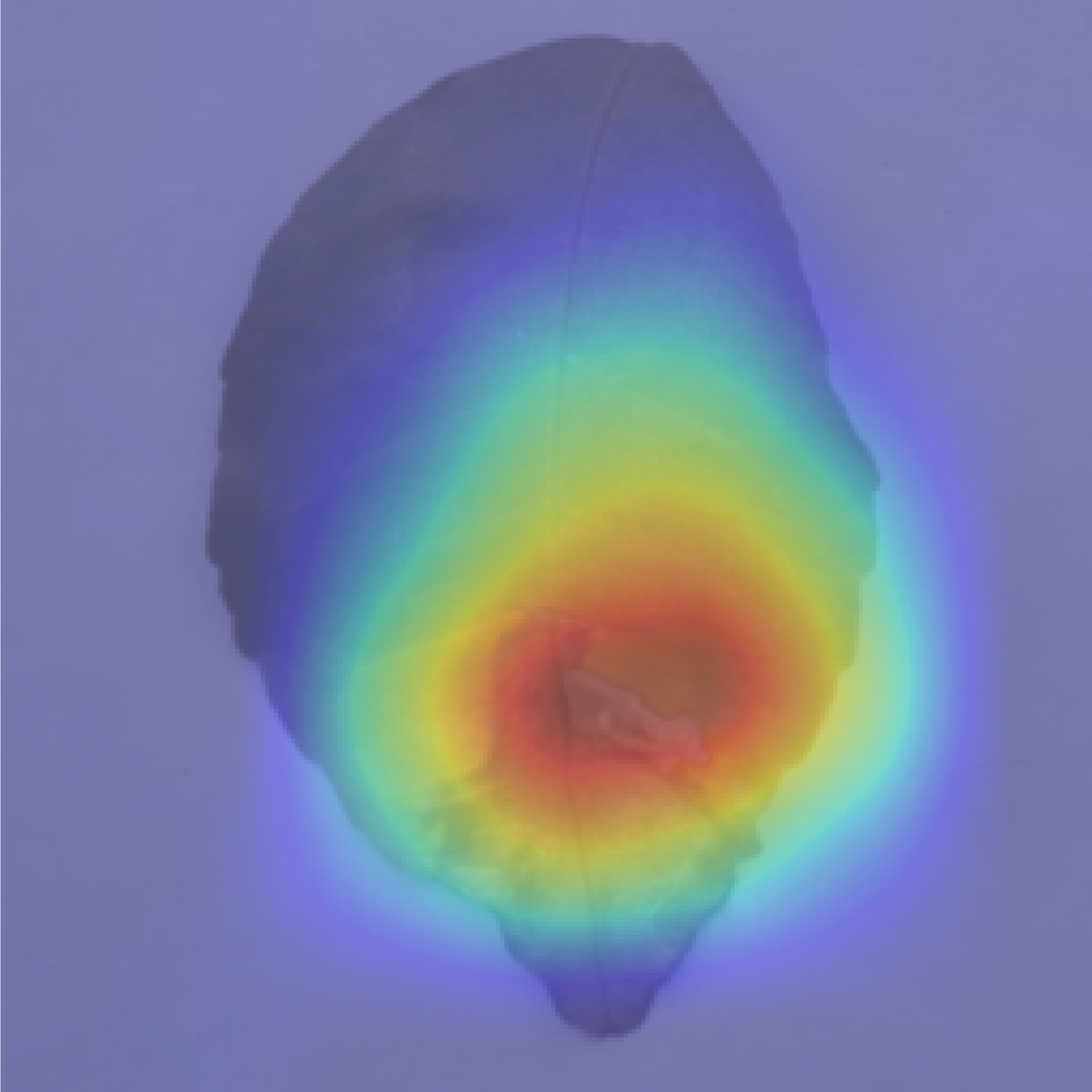}
        \caption{SSM}
        \label{fig:c}
    \end{subfigure}

    \vspace{0.3cm}

    \begin{subfigure}{0.12\textwidth}
        \centering
        \includegraphics[width=\linewidth]{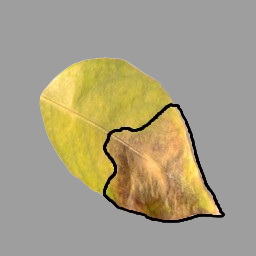}
        \caption{Input}
        \label{fig:d}
    \end{subfigure}
    \hspace{1.5em}
    \begin{subfigure}{0.12\textwidth}
        \centering
        \includegraphics[width=\linewidth]{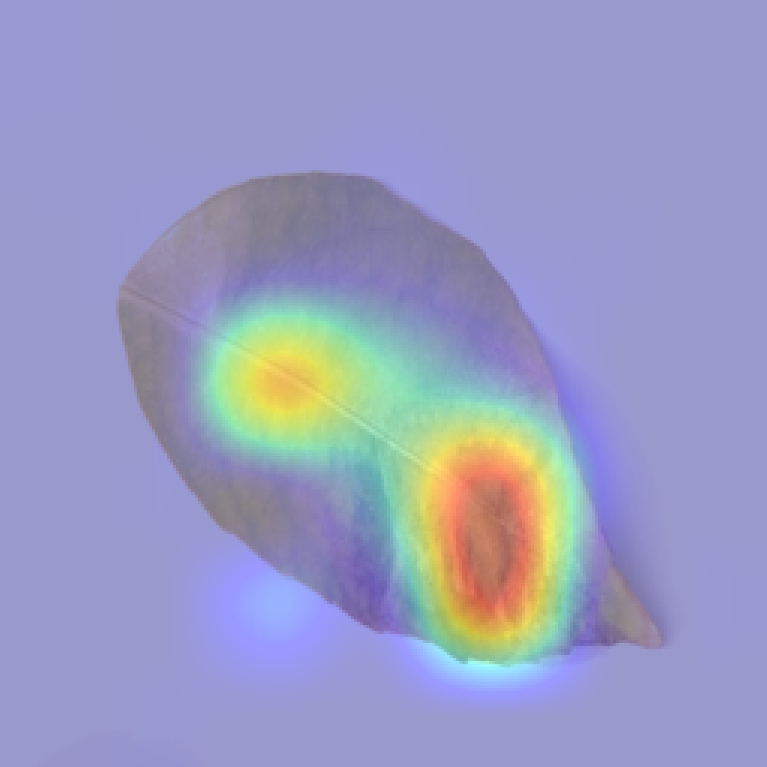}
        \caption{DINO}
        \label{fig:e}
    \end{subfigure}
    \hspace{1.5em}
    \begin{subfigure}{0.12\textwidth}
        \centering
        \includegraphics[width=\linewidth]{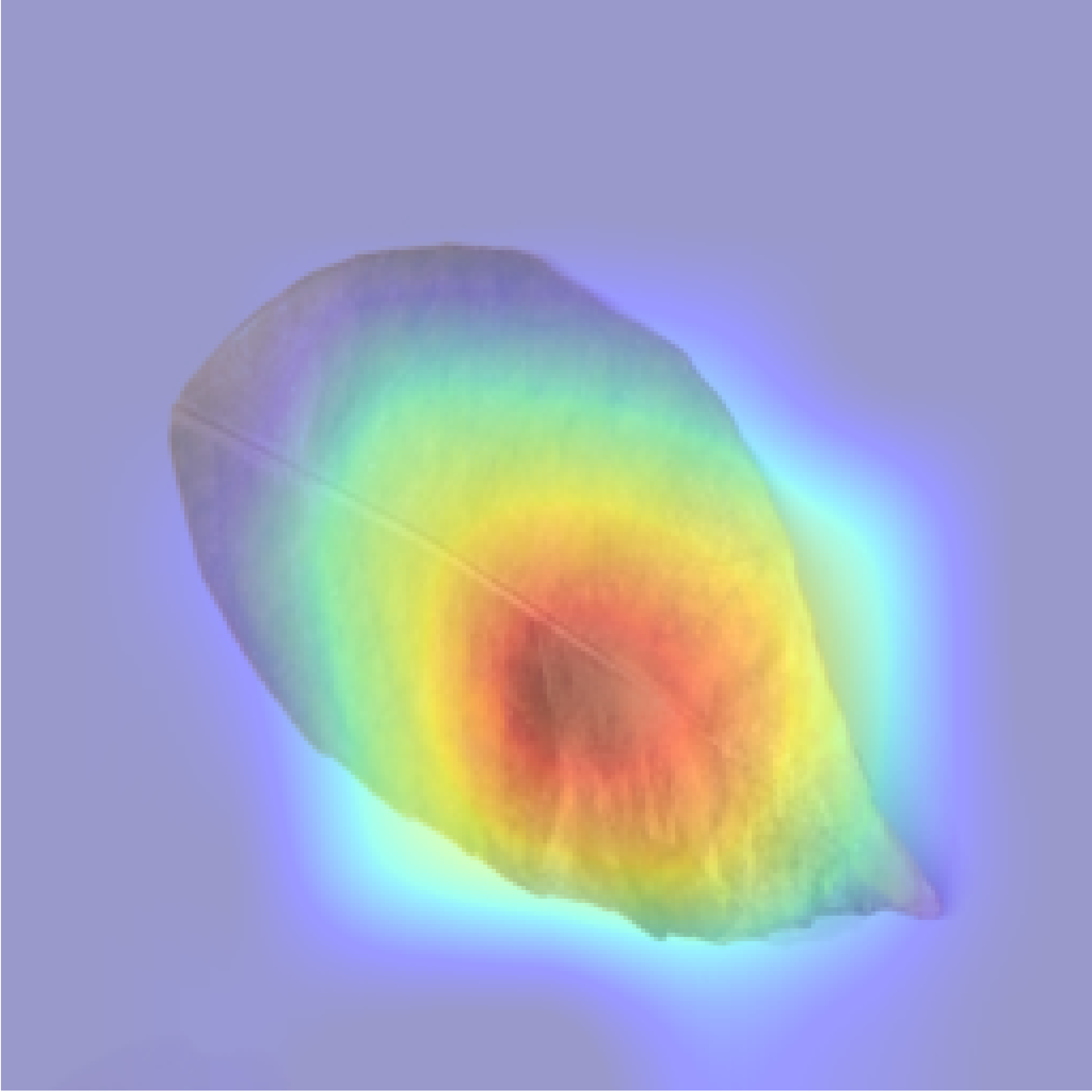}
        \caption{SSM}
        \label{fig:f}
    \end{subfigure}

    \caption{Grad-CAM visualisations for two leaf samples. Input images (a, d) show the diseased regions. DINO (b, e) exhibits diffuse or background-biased attention, whereas StateSpace-SSL (c, f) provides better localisation.}
    \label{fig:visual_compare}
\end{figure}

This trend is supported by the training-time and memory statistics in Table \ref{tab:timing_compare} and the per-epoch comparison in Figure \ref{fig:time_per_epoch}. StateSpace-SSL converges in only 9 hours of pretraining while using just 8.7 GB of VRAM, whereas transformer-based approaches typically require 17-30 hours and substantially higher memory. The per-epoch comparison in Figure \ref{fig:time_per_epoch} highlights this contrast: StateSpace-SSL completes each epoch in approximately 1.8 minutes, compared with 3.4-4.8 minutes for DeiT \cite{touvron2021training}, DINO \cite{caron2021emerging}, and Hybrid ViT \cite{dosovitskiy2020image}, and around 6 minutes for MAE \cite{he2022masked}. The consistently lower iteration cost reflects the efficiency of the state-space update mechanism, which avoids the quadratic bottlenecks that arise in larger transformer architectures.
The qualitative visualisation in Figure \ref{fig:visual_compare} further supports these quantitative findings. Using Grad-CAM on two representative sample images, we compare the spatial attention patterns produced by DINO \cite{caron2021emerging}, the strongest transformer baseline in our experiments, and the proposed StateSpace-SSL. While DINO \cite{caron2021emerging} frequently generates diffuse or background-oriented activations, particularly around leaf boundaries or non-informative regions, StateSpace-SSL produces more compact and lesion-centred responses. This behaviour suggests that the state-space encoder aggregates contextual cues more coherently, enabling it to focus on relevant disease structures. The improved localisation aligns with the accuracy gains reported earlier and illustrates how the linear state-space formulation yields representations that are both discriminative and robust to background noise typical of field-captured plant disease imagery.

Taken together, these results show that StateSpace-SSL offers a favourable balance of accuracy, robustness, and computational efficiency. By replacing attention-heavy architectures with a lightweight state-space formulation, the method provides a more effective inductive bias for plant disease imagery, especially under field conditions where background variability can be substantial. The combination of strong generalisation and efficient training dynamics makes StateSpace-SSL well-suited for large-scale pretraining pipelines as well as practical deployment scenarios that require models to be both reliable and resource-efficient.

\subsection{Discussion and Limitation}

The results indicate that linear state-space modelling is well-suited for plant disease detection, where lesion patterns must be captured reliably under varying field conditions. StateSpace-SSL provides stable representations across both controlled and real-world datasets, suggesting that the implicit long-range mixing of state-space layers aligns well with the spatial structure of plant pathology symptoms. The strong performance on PlantDoc further demonstrates that the model handles background variability and illumination changes more effectively than transformer-based baselines, which often over-attend to non-diagnostic regions. Importantly, these benefits are achieved with a substantially lighter architecture, confirming that strong disease discrimination does not require the high computational complexity typical of transformer models. Although the method is effective overall, it may still face challenges when disease symptoms are extremely subtle or occupy only small regions of the leaf, where additional spatial precision may be required. A natural extension is to incorporate cross-scale mechanisms or auxiliary spectral cues to improve sensitivity to these fine-grained patterns.

\section{Conclusion}
This paper introduced StateSpace-SSL, a linear-time self-supervised learning framework tailored to plant disease detection. By leveraging a Vision Mamba encoder and a prototype-based teacher–student alignment strategy, the method learns robust disease-relevant representations while avoiding the quadratic costs of transformer attention. Experiments on PlantVillage, PlantDoc, and Citrus show that StateSpace-SSL achieves superior accuracy, stronger generalisation in real-field conditions, and substantially lower memory requirements and training time than SSL baselines. The qualitative results further confirm that state-space dynamics yield more lesion-focused responses, enabling reliable detection even in complex field backgrounds. 

\bibliography{aaai2026}

\end{document}